\documentclass[journal]{IEEEtran}
\usepackage{subfig}

\usepackage[pagebackref]{hyperref}
\usepackage{diagbox}

\usepackage{booktabs}
\hypersetup{
    colorlinks=true,
    linkcolor=blue,
    citecolor=blue,
    urlcolor=blue
}
\usepackage{graphicx}
\usepackage{multirow}

\ifCLASSINFOpdf
\else
\fi
\usepackage{amsmath}
\usepackage{amssymb}
\usepackage{threeparttable}

\begin{document}
\title{FingerSplat: \\Contactless Fingerprint 3D Reconstruction and Generation based on 3D Gaussian Splatting}

\author{Yuwei~Jia\textsuperscript{\textdagger},~\IEEEmembership{Student~Member,~IEEE,}
        Yutang~Lu\textsuperscript{\textdagger},
        Zhe Cui*,~\IEEEmembership{Member,~IEEE},
        Fei Su,~\IEEEmembership{Member,~IEEE},% <-this % stops a space
\thanks{\textsuperscript{\textdagger}Yuwei Jia and Yutang Lu contributed equally to this work.}
\thanks{*Corresponding author: Zhe Cui (email: cuizhe@bupt.edu.cn).}%
\thanks{Y. Jia, Y. Lu, Z. Cui, F Su are with the Beijing Key Laboratory of Network System and Network Culture, Beijing University of Posts and Telecommunications, Beijing, China. (cuizhe@bupt.edu.cn)}
}% <-this % stops a space
% \thanks{J. Doe and J. Doe are with Anonymous University.}% <-this % stops a space
% \thanks{Manuscript received April 19, 2005; revised August 26, 2015.}}


\markboth{Journal of \LaTeX\ Class Files,~Vol.~14, No.~8, August~2015}%
{Shell \MakeLowercase{\textit{et al.}}: Bare Demo of IEEEtran.cls for IEEE Journals}



\maketitle

\begin{abstract}
Researchers have conducted many pioneer researches on contactless fingerprints, yet the performance of contactless fingerprint recognition still lags behind contact-based methods primary due to the insufficient contactless fingerprint data with pose variations and lack of the usage of implicit 3D fingerprint representations. In this paper, we introduce a novel contactless fingerprint 3D registration, reconstruction and generation framework by integrating 3D Gaussian Splatting, with the goal of offering a new paradigm for contactless fingerprint recognition that integrates 3D fingerprint reconstruction and generation. To our knowledge, this is the first work to apply 3D Gaussian Splatting to the field of fingerprint recognition, and the first to achieve effective 3D registration and complete reconstruction of contactless fingerprints with sparse input images and without requiring camera parameters information. Experiments on 3D fingerprint registration, reconstruction, and generation prove that our method can accurately align and reconstruct 3D fingerprints from 2D images, and sequentially generates high-quality contactless fingerprints from 3D model, thus increasing the performances for contactless fingerprint recognition. 
% Our approach can synthesize new viewpoints of contactless fingerprint images, thereby augmenting existing contactless fingerprint datasets. At the same time, our method performs direct registration of contactless fingerprints in 3D space, offering a novel perspective for both contactless fingerprint registration and template construction. 

% Moreover, by directly registering contactless fingerprints in 3D space, our method provides a new strategy for contactless fingerprint alignment and template construction.
\end{abstract}


% \fi
\IEEEpeerreviewmaketitle

\section{Introduction}

% \begin{figure}[!t]
% \centering
% \includegraphics[width=2.5in]{imgs/overview.pdf}
% where an .eps filename suffix will be assumed under latex, 
% and a .pdf suffix will be assumed for pdflatex; or what has been declared
% via \DeclareGraphicsExtensions.
% \caption{Simulation results for the network.}
% \label{fig_sim}
% \end{figure}
% \begin{figure}
% \centering
% \centerline{\includegraphics[width=\linewidth]{imgs/first_image.pdf}}
% \caption{The proposed method can reconstruct 3D Gaussian Splatting from 3 Contactless Fingerprint Images.
% \label{fig:first_image}
% \end{figure}

\begin{figure}[!bt]
\centering
\centerline{\includegraphics[width=\linewidth]{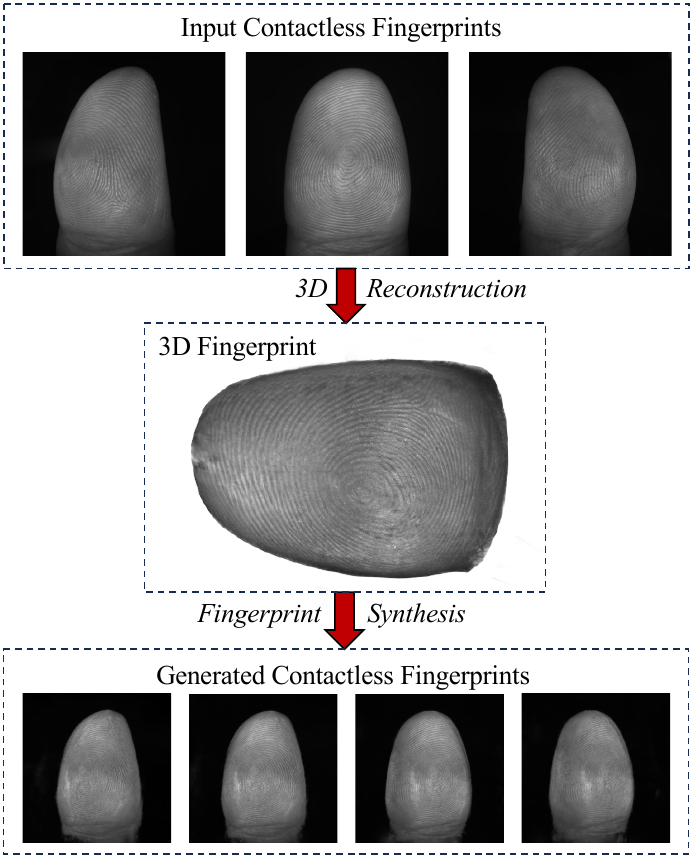}}
\caption{The proposed method can reconstruct 3D Gaussian Splatting from 3 contactless fingerprint images and then generate contactless fingerprint images.}
\label{fig:overview}
\vspace{-15pt}
\end{figure}

\IEEEPARstart{F}{ingerprint} is one of the most widely used biometric modalities for identity authentication. Compared with conventional contact-based fingerprints, contactless fingerprints provide a different touchless acquisition approach and lead to specific researches on the contactless fingerprint. In recent years, many contactless fingerprint recognition methods have been proposed\cite{tan2020towards}\cite{dong2023syn}\cite{shi2022towards}\cite{yin2019contactless}\cite{grosz2022contact}\cite{yin2021fingerprint}\cite{dong2025bridging}. However, due to the issues such as imaging quality, lighting conditions, and finger pose variations, the recognition performances of contactless fingerprints still struggle to achieve practically useful recognition accuracy compared with contact-based fingerprints, especially contactless fingerprints with large pose variation and small overlapping area. Several methods \cite{tan2020towards}\cite{cui2023monocular} furtherly introduce the 3D model into contactless fingerprint recognition to solve pose issues, but their methods often need additional information for 3D fingerprint reconstruction, which introduces extra burdens. 

Another main limitation for contactless fingerprint method is the insufficient training data, primarily due to specific acquisition equipment and biometric privacy issues, which further limits the research and application of contactless fingerprints. For data issues, synthetic methods \cite{grosz2022spoofgan}\cite{engelsma2022printsgan}\cite{grosz2024universal}\cite{dong2023syn}\cite{dong2025bridging} are introduced to enlarge the contactless fingerprint datasets. Some synthetic methods \cite{grosz2022spoofgan}\cite{engelsma2022printsgan}\cite{grosz2024universal} are 2D-image based, while several methods \cite{dong2023syn}\cite{dong2025bridging} are 3D-fingerprint synthetic methods. But the synthetic methods are often data-dependent, and can not generate high-quality and high-fidelity fingerprint images.

Recently, the advent of deep learning in fingerprint 3D reconstruction \cite{cui2023monocular} and fingerprint alignment \cite{guan2024phase}\cite{jia2025single} has encouraged us to incorporate these achievements into fingerprint 3D reconstruction. Our method further demonstrates that, it is feasible to directly reconstruct and render the fingerprint in 3D space from sparse contactless fingerprint images without camera parameters, and that the resulting 3D fingerprint can be used to synthesize new viewpoints of contactless fingerprint images, thereby augmenting contactless fingerprint datasets to eventually improve recognition performances.

\begin{figure*}[!bt]
\centering
\centerline{\includegraphics[width=\linewidth]{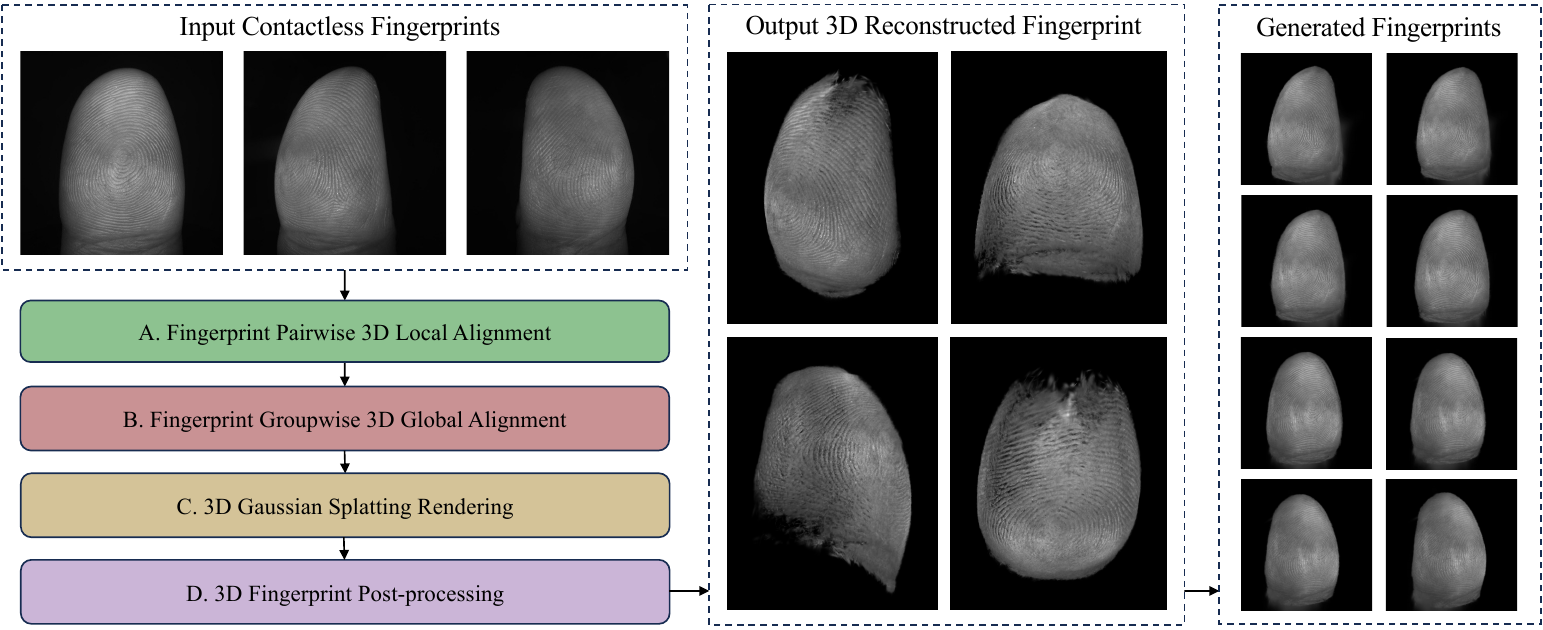}}
\caption{Overview structure of the proposed FingerSplat}
\label{fig:overview}
\vspace{-10pt}
\end{figure*}

% To improve these issues, we propose a 3D reconstruction framework for contactless fingerprints that can effectively stitch together contactless fingerprints with large pose variations and small overlap areas in 3D space, then synthesize images from arbitrary viewpoints to augment training data, effectively addressing this problem.

\subsection{Related Works}

\textit{1) Contactless Fingerprint Recognition: } A major challenge for contactless fingerprint recognition methods is the difficulty in effectively matching contactless fingerprints with large pose variance. To address this issue, Tan et al. \cite{tan2020towards}, Cui et al. \cite{cui2023monocular}, Grosz et al. \cite{grosz2022contact}, and Yin et al.\cite{yin2019contactless}
focus mainly on correcting perspective distortions across different views of contactless fingerprints.
Tan et al. \cite{tan2020towards} normalize the orientation of contactless fingerprints,
while other methods \cite{cui2023monocular}\cite{grosz2022contact}\cite{yin2019contactless} unify ridge frequency—whether in 3D
or 2D space-to reduce distortion. Dong et al. \cite{dong2023syn} adopt the network-based method to correct fingerprint distortion, and Shi et al. \cite{shi2022towards} attempt to learn unique contactless fingerprint features 
via graph neural networks. Although these methods address view variations through decreasing perspective distortion at some level, the pose variation issue is still not solved yet, as these methods match contactless fingerprints from different views in an independent way, neglecting the complementary information across different finger views to form a complete 3D fingerprint, which is beneficial for recognition.

\textit{2) Contactless Fingerprint 3D Reconstruction: } A common approach for 3D fingerprint reconstruction is to perform multi-view 3D reconstruction when the camera’s intrinsic and extrinsic parameters are known \cite{liu20143d}\cite{parziale2006surround}\cite{labati2015toward}. Methods such as shape from shading \cite{kumar2013towards}\cite{lin2017tetrahedron} and shape from focus \cite{abramovich2010mobile} can reconstruct 3D fingerprints, but they require expensive and complex acquisition equipment. There are also 3D fingerprint acquisition methods based on structured light \cite{wang2010data}\cite{wang20233d}, ultrasound imaging \cite{baradarani2013resonance}, and laser \cite{galbally2017full}, but they also rely on dedicated acquisition devices. Some previous work has tried to mosaic contactless fingerprint images \cite{alkhathami2014mosaic}\cite{liu2013touchless}\cite{liu2020advanced} to make a complete contactless fingerprint, but these methods have not been tested on publicly available contactless fingerprint datasets without camera pose or true depth. The mosaicked results heavily depend on the specifics of the camera capture setup, which is not practical.
% In contrast, our method requires no extra finger data for pretraining; we can directly synthesize 3D fingerprints from the three views available in the UWA dataset.

% In practical applications, fingerprint recognition is not limited to pairwise image matching but is often carried out via retrieval, where the template typically contains the entire distal phalanx fingerprint. Such a complete template is difficult to obtain in existing contactless fingerprint capture processes. In contact-based fingerprint retrieval, we often use rolled fingerprints, which cover a large surface area and effectively resolve the problem of small overlap between the input and stored prints. When only a small-area fingerprint can be acquired—such as a partial fingerprint on a smartphone—a common practice is to stitch partial fingerprints together to form a larger template. 
% To stitch contactless fingerprint images, one idea is to first predict their depth and then unwrap them, followed by using a registration algorithm designed for contact fingerprints to perform the stitching. This method relies on binarizing the fingerprint images; otherwise, it is hard to determine the optimal seam line. Current fingerprint binarization methods \cite{fingernet2017}\cite{verifinger}\cite{zhu2023fingergan} are mainly designed for contact fingerprints and yield unsatisfactory results when applied to contactless fingerprints. 

\textit{3) Fingerprint Synthesis: } Research on fingerprint synthesis has a long history, including early work that used traditional approaches to generate fingerprints \cite{cappelli2003synthetic}\cite{cappelli2004improved}, as well as studies based on GANs \cite{grosz2022spoofgan}\cite{engelsma2022printsgan}. More recent research has adopted state-of-the-art diffusion models to synthesize fingerprints and even contactless fingerprint images \cite{grosz2024universal}. However, these fingerprint synthesis methods can only generate two-dimensional fingerprint images. In recent years, several methods \cite{priesnitz2022syncolfinger}\cite{dong2023syn}\cite{dong2025bridging}\cite{grosz2024universal} have been proposed to synthesize contactless fingerprint data. Priesnitz et al. \cite{priesnitz2022syncolfinger} mapped fingerprint textures onto skin images in 2D space, but their approach cannot generate fingerprints from different viewpoints. Dong et al. \cite{dong2023syn} addressed this limitation by reprojecting textures onto a 3D Bézier model. Nevertheless, these methods cannot synthesize new views from existing contactless fingerprints, and the quality of the generated images remains limited.  Recently, some work has introduced implicit 3D model representations into finger biometrics \cite{xu2024improving}. This work builds its own finger‐video dataset and are successfully trained a NeRF to reconstruct 3D fingers and generate contactless fingerprints. However, their experiments on the UWA dataset \cite{zhou2014benchmark} rely on pretraining with dense multi-view inputs, which heavily relies on the large amount of additional fingerprint data.

\subsection{Objective and Key Contributions}

Based on the recent achievements of 3D Gaussian Splatting \cite{kerbl20233d} and 3D reconstruction \cite{wang2024dust3r}, we propose a direct 3D fingerprint reconstruction method from contactless fingerprint images captured at different angles of the same fingerprint. The core idea is to first align a pair of contactless fingerprints in 3D space using 3D correspondences matching method \cite{wang2024dust3r}. Then, based on pairwise correspondences, we estimate global camera parameters to complete the initial 3D point‐cloud reconstruction of the fingerprint. Next, we refine both the point cloud and camera poses using 3D Gaussian Splatting. Finally, we perform post-processing of the 3D Gaussian scene using SAM segmentation \cite{hu2024semantic}, resulting in a 3D Gaussian representation of the fingerprint that can be used for template construction or rendering new contactless fingerprints (see Fig. \ref{fig:overview}). 

Our approach is able to preserve the original fingerprint image information while achieving high-quality 3D fingerprint results, avoiding the information loss during 3D-2D transition. Our method can further reconstruct and stitch a 3D fingerprint using 2D contactless fingerprint images—captured from poses with large variations—without any annotated camera parameters, demonstrating strong generalizability. 
% Our approach is mainly inspired by DUSt3R \cite{wang2024dust3r}\cite{leroy2024grounding}, LoFTR \cite{sun2021loftr}\cite{chen2022aspanformer}, 3D Gaussian Splatting \cite{kerbl20233d}\cite{fan2024instantsplat}, and SAM \cite{kirillov2023segment}\cite{cen2025segment}\cite{hu2024semantic}, among other methods in computer vision and graphics. 

In general, our work has four contributions:
% \vspace{0pt} % 减少图表下方的间距

\begin{itemize}

\item[1)] We propose a fingerprint 3D reconstruction framework that can effectively reconstruct high‑quality 3D models of contactless fingerprints from sparse‑view 2D inputs (e.g. only three viewpoints).
\item[2)] We are the first to introduce 3D Gaussian Splatting into the domain of fingerprint recognition, demonstrating that it enables high-quality and photorealistic generation of multi-view contactless fingerprints.
%\cite{kumar2013towards}\cite{liu2015study}\cite{lin2017tetrahedron} that requires specific designed equipment and multiple images to acquire a 3D shape, our method only needs a single contactless fingerprint and do not require additional information assistance.
\item[3)] We innovatively register contactless fingerprints in 3D space, achieving more accurate alignment than traditional 2D registration methods. Moreover, our registration approach adheres better to physical principles.
\item[4)] Experimental results demonstrate that our method can align, reconstruct, and synthesize contactless fingerprints in 3D space, and generates high-quality contactless fingerprint images. Therefore, our method can effectively broadens the variety of contactless fingerprint images and improve the recognition performance for contactless fingerprints.

\end{itemize}

\section{Method}

This section details the methodology for reconstructing and synthesizing contactless 3D fingerprints of our method, including: A. Fingerprint Pairwise 3D Local Alignment, B. Fingerprint Groupwise 3D Global Alignment, C. 3D Gaussian Splatting Rendering. D. 3D Fingerprint Post-processing. The proposed method is able to generate 3D as well as 2D fingerprint through 3D fingerprint reconstruction and rendering by 3D Gaussian Splatting. Moreover, the proposed method is able to operate effectively even under unknown finger pose scenes and sparse input fingerprint images, which is especially critical and surpasses previous contactless fingerprint methods.

\subsection{Fingerprint Pairwise 3D Local Alignment}

\begin{figure}[!bt]
\centering
\centerline{\includegraphics[width=\linewidth]{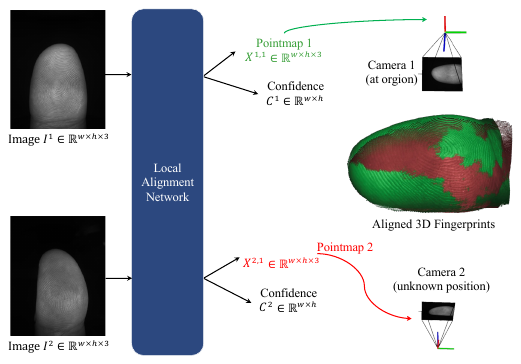}}
\caption{Fingerprint Pairwise Local Alignment}
\label{fig:network}
\vspace{-10pt}
\end{figure}

In our framework, the first step is to align contactless fingerprint images in a 3D space to generate complete 3D fingerprint data. This 3D alignment is two-stage, we first perform fingerprint alignment in pairs, and then perform global optimization based on these pairwise alignment results to generate a full 3D fingerprint. Unlike previous fingeprint registration methods that match fingerprint features in 2D space, we directly match fingerprints utilizing 3D point cloud registration \cite{wang2024dust3r}. Since it is very expensive to manually annotate point correspondences on contactless fingerprints in 3D space, we adopt the original pretrained model of \cite{wang2024dust3r} in the first step of local alignment, and finetune the result in the following global alignment. 

As shown in Fig. \ref{fig:network}, the core of our pairwise local alignment step is an end-to-end deep network that directly regresses 3D fingerprint pointmap from unconstrained fingerprint images, which can be viewed as the position of each point in a 2D image within three-dimensional space. These pointmaps are representations that contain dense 3D geometric information, and all pointmaps are expressed in the same reference view, thereby implicitly encoding the geometric relationships between views. It is worth noting that our method directly uses a pre‑trained model to obtain the depth of contactless fingerprints, achieving performance comparable to previous monocular fingerprint depth prediction approaches \cite{cui2023monocular}.

% The entire network is trained using a unified, confidence-aware regression loss function, \(\mathcal{L}_{\text{conf}}\). The objective of this loss function is to minimize the 3D Euclidean distance between the predicted pointmap and the ground-truth pointmap after scale alignment.
% \begin{equation}
% \mathcal{L}_{\text{conf}} = \sum_{v \in \{1,2\}} \sum_{i \in \mathcal{D}^v} C_{i}^{v,1} \ell_{\text{reg}}(v,i) - \alpha \log C_{i}^{v,1} \quad
% \end{equation}
% Here, \(\ell_{\text{reg}}(v,i)\) is the 3D regression loss for a single point, \(C_{i}^{v,1}\) is the confidence score predicted by the network for each point, and \(\alpha\) is a regularization hyperparameter.

% From the dense pointmaps output by the network, various key components of the traditional 3D vision pipeline can be easily derived.

% \subsubsection*{Depth Estimation}
% In our framework, depth estimation is a direct product from the pointmap regression task. For any given image, its corresponding depth map is simply the Z-coordinate of each 3D point in the predicted pointmap.

% Theoretically, the relationship between a pointmap \(X\), a depth map \(D\), and camera intrinsics \(K\) can be expressed as:
% \begin{equation}
% X_{i,j}=K^{-1}D_{i,j}[i,j,1]^{\dagger} \quad 
% \end{equation}
% In our method, we regress the pointmap \(X\) directly, rather than relying on pre-estimated \(D\) and \(K\).

% \subsection{Estimate global position and camera parameter of each view}
\subsection{Fingerprint Groupwise 3D Global Alignment}
After local alignment, global alignment is performed to obtain the global positions and camera parameters of all fingerprint images through joint optimization of the pointmaps. By this step, we can obtain a complete 3D point cloud of the finger, which is then used for the subsequent contactless fingerprint synthesis rendering via 3D Gaussian Splatting.
% and local feature matching result by SFRNet..
\begin{itemize}
% \item \textbf{2D Local Matching}
% We SFRNet's which is trained on NIST302 \cite{sd302} to get the 2D local feature matching result.

\item \textbf{Camera and Pose Initialization}
\begin{itemize}
    \item \textbf{Camera Initialization}: Since the pointmap is expressed in the camera's coordinate frame, camera intrinsics (such as focal length) can be recovered by optimizing the reprojection error of the pointmap onto the image plane .
    \item \textbf{Pose Initialization}: The relative pose between two cameras can be recovered by Procrustes alignment to compare two pointmaps (e.g., \(X^{1,1}\) and \(X^{2,2}\)), which yields the scaled relative pose \(P^* = \sigma^* [R^* | t^*]\) . Its optimization objective is:
    \begin{equation}
    P^* = \arg\min_{\sigma, R, t} \sum_{i} C_{i}^{1,1} C_{i}^{2,2} \left\| \sigma (R X_{i}^{1,1} + t) - X_{i}^{2,2} \right\|^2 \quad 
    \end{equation}.
    Where $C$ represents the confidence of the corresponding point $X$.
\end{itemize}

\item \textbf{3D Global Alignment}

We adopt a Global Alignment optimization step which aims to align all pairwise predicted pointmaps into a globally consistent 3D space, forming a complete model of the scene. Its optimization problem is defined as minimizing the 3D projection error between each pairwise prediction and the final global model:
\begin{equation}
\chi^{*}= \arg\min_{\chi,P,\sigma}\sum_{e\in\mathcal{E}}\sum_{v\in e}\sum_{i=1}^{HW}C_{i}^{v,e}||\chi_{i}^{v}-\sigma_{e}P_{e}X_{i}^{v,e}|| \quad 
\end{equation}
Here, \(e=(n,m)\) represents a pair of images, \(X_i^{v,e}\) is the 3D point predicted in the reference frame of pair \(e\), \(P_e\), \(C_i^{v,e}\) is the corresponding confidence and \(\sigma_e\) are the pose and scale parameters for that pair, and \(\chi_i^v\) is the global 3D point to be optimized.
\end{itemize}

\subsection{3D Gaussian Splatting Rendering}
\begin{figure}[!bt]
\centering
\centerline{\includegraphics[width=\linewidth]{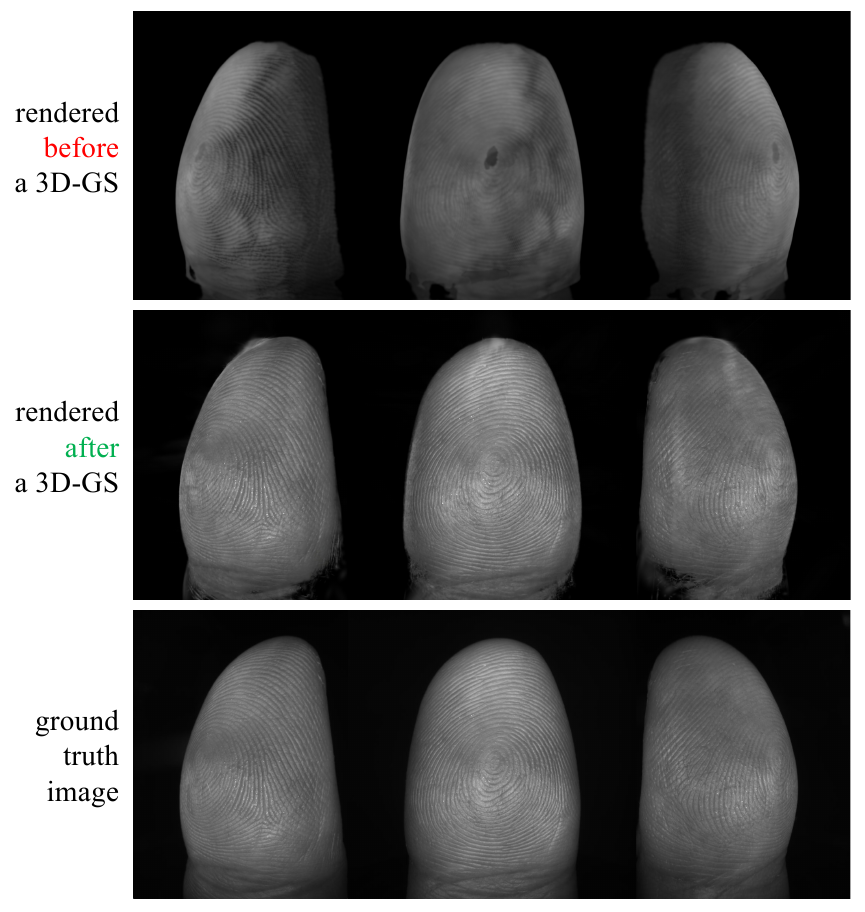}}
\caption{Comparison of rendered 3D fingerprints with or without 3D Gaussian Splatting.}
\label{fig:3dgs_after}
\vspace{-5pt}
\end{figure}

Due to illumination effects, directly stitching a 3D fingerprint from point‑maps often shows obvious seams, which introduce substantial noise to the fingerprint’s surface texture. Therefore, we incorporate 3D Gaussian Splatting to further optimize the 3D model with ground‑truth viewpoints to remove artifacts and generate fingerprints with higher quality, as shown in Fig. \ref{fig:3dgs_after}.

3D Gaussian Splatting is an explicit 3D scene representation technique that models scenes using a collection of 3D Gaussians. Each 3D Gaussian is defined by a mean vector $\mathbf{x} \in \mathbb R^3$, an opacity $\alpha \in \mathbb R$, and a covariance matrix $\Sigma \in \mathbb R^{3 \times 3}$:
\begin{equation}
G(\mathbf{p}, \alpha, \Sigma) = \alpha \exp \left( -\frac{1}{2} (\mathbf{p} - \mathbf{x})^T \Sigma^{-1} (\mathbf{p} - \mathbf{x}) \right)
\end{equation}
To handle view-direction-dependent effects, spherical harmonic (SH) coefficients are attached to each Gaussian, and the color is rendered using the view-dependent color and opacity. However, 3D fingerprints typically do not require complex lighting modeling. Moreover, in certain datasets, the data acquisition is not performed by capturing the same scene from different viewpoints, but rather by using a fixed camera position while varying the finger pose. Therefore, when processing colors, we only retain the base color representations without the addition for spherical harmonics. This simplifies the calculation process while ensuring the quality of the rendering results.

The training loss function of 3DGS rendering is defined as:
\begin{equation}
\mathcal{L} = (1-\lambda_{\text{SSIM}})\mathcal{L}_1 + \lambda_{\text{SSIM}} \mathcal{L}_{\text{SSIM}}
\end{equation}

where:
\begin{itemize}
    \item \(\mathcal{L}_1\) is the photometric (L1) loss:
    \begin{equation}
    \mathcal{L}_1 = \frac{1}{N}\sum_{i=1}^{N} |G_i - R_i|
    \end{equation}
    Here, \(G_i\) and \(R_i\) are the pixel values of the ground truth and rendered images, respectively, and \(N\) is the total number of pixels.

    \item \(\mathcal{L}_{\text{SSIM}}\) is the D-SSIM loss:
    \begin{equation}
    \mathcal{L}_{\text{SSIM}} = 1 - \text{SSIM}(G, R)
    \end{equation}
    SSIM is the Structural Similarity Index Measure, which considers luminance, contrast, and structural information.

    \item \(\lambda_{\text{SSIM}}\) is the weight for the D-SSIM term, typically set to 0.2.
\end{itemize}

\subsection{3D Fingerprint Post-processing}

\begin{figure}[!bt]
\centering
\centerline{\includegraphics[width=\linewidth]{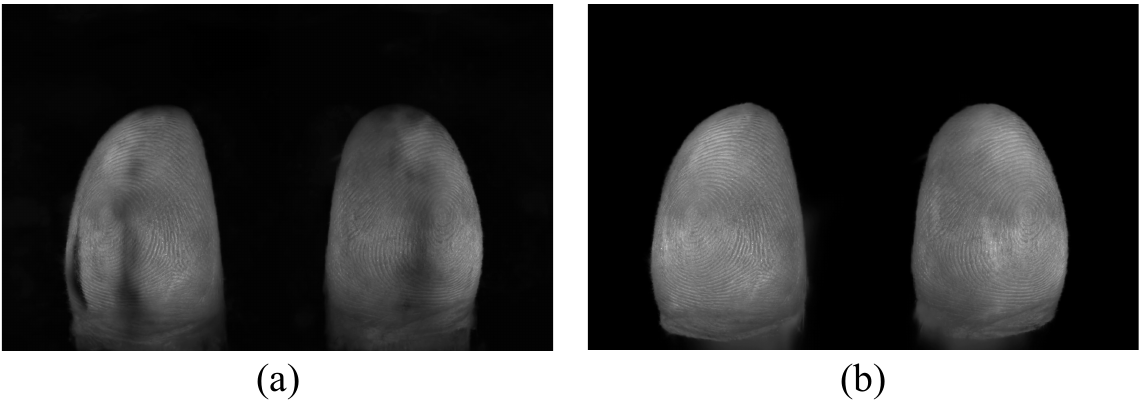}}
\caption{Comparison of rendering point clouds without or with 3D segmentation post-processing. (a) Without post-processing, (b) With post-processing.}
\label{fig:sagd}
\vspace{-10pt}
\end{figure}

The contactless fingerprint rendered by 3D-GS \cite{kerbl20233d} is able to deliver high-fidelity, real-time novel-view synthesis. However, the direct output of 3D-GS uses boundary Gaussians to straddle multiple objects, leading to rough segmentation edges, which needs post-processing. Since the main issue of direct 3D-GS output is the unclear boundary, we conduct a SAGD \cite{hu2024semantic} based 3D fingerprint segmentation step by combining multi-view mask generation, Gaussian decomposition, and label voting. As shown in Fig. \ref{fig:sagd}, when rendering with 3D Gaussian Splatting (3D‑GS) using a black background, a large amount of noise is introduced, which degrades the performance of novel view synthesis. The segmentation post-processing outputs a clean boundary result.

\section{Experiment}

Our method is able to reconstruct, register, and synthesize contactless fingerprints in 3D space, and the following experiments examine our method. First, we test the depth prediction capability of our proposed method for testing 3D reconstruction. Sceond, we test our contactless fingerprint 3D alignment procedure as a fingerprint registration task and conduct comparative experiments. Then, we evaluate the quality of our synthesized contactless fingerprints using image metrics and fingerprint quality assessment metrics. Moreover, we verify that our synthetic fingerprints can be used for fingerprint recognition through matching experiments between synthesized and original fingerprints. Finally, we apply the synthesized new fingerprints to practical applications, including training fingerprint classification networks for fingerprint matching and using them as templates for fingerprint retrieval.

\subsection{Datasets and Experiment Details}

Most of the available contactless fingerprint datasets \cite{lin2017tetrahedron}\cite{lin2018matching}\cite{birajadar2019towards}\cite{sankaran2015smartphone}\cite{malhotra2020matching}\cite{grosz2022contact} lack significant pose variations, making it difficult to complete 3D fingerprints. The most suitable datasets for our experiments are UWA Benchmark 3D/2D Fingerprint Database \cite{zhou2014benchmark} and CFPose Database \cite{tan2020towards}, so we conduct our experiments primarily on these two datasets. In addition, we employed the PolyU 3D+ Database \cite{lin2017tetrahedron}, which provides depth ground truth, to conduct depth prediction experiments.

\textbf{UWA Benchmark 3D/2D Fingerprint Database} contains 8,958 contactless fingerprints with pose variations. This dataset uses sensors to capture contactless fingerprints from three viewpoints simultaneously, eliminating variations in light source and finger position. We use the three contactless fingerprints from the first capture for all subsequent experiments. Meanwhile, we conducted depth prediction experiments using the PolyU 3D+ Database, which contains depth ground truth.

\textbf{CFPose Database} contains 1,400 contactless fingerprints with random pose variations. 
% Although this dataset performs poorly by baseline, our improvements effectively enhance fingerprint synthesis results. 
For this database, we predict the yaw angle of the fingerprints by \cite{tan2020towards} and select the image with the smallest angle and the two with the largest angle from ten different viewpoints to synthesize 3D fingerprints for all subsequent experiments. Since the data collection protocol of this dataset is different from UWA, we preprocess the dataset by upright‑rotation using the method of Cui et al. \cite{cui2023monocular}. We also rescale its image width to match the UWA dataset (1024 pixels), and cropp out the black background at the top, retaining only a height of 1280 pixels.

\textbf{PolyU 3D+ Database} \cite{lin2017tetrahedron} We utilize the 2,016 ground truth depth maps and corresponding images from the first session for depth estimation experiments.

Since the 3D Gaussian Splatting is able to generate any views of fingerprints, we extract 12 frames at equal intervals between the left, front, and right viewpoints in both datasets as our synthesized fingerprint images for subsequent experiments.

We generally adopt the default parameter settings of InstantSplat \cite{fan2024instantsplat} and SAGD \cite{hu2024semantic}, with the sole exception of setting SH\_DEGREE = 0. 

\subsection{3D Reconstruction Accuracy}

To validate the effectiveness of our method on 3D fingerprint reconstruction, we perform depth prediction experiments on the PolyU 3D+ Database \cite{lin2017tetrahedron} with real depth ground truth. Since the fingerprint alignment step requires at least two input images while most current fingerprint depth estimation methods are single-input, we feed two identical images during testing to obtain depth, effectively converting our method into a monocular depth estimation approach for comparison with existing monocular method \cite{cui2023monocular}. As shown in Table \ref{depth}, our method significantly outperforms that of Cui et al. \cite{cui2023monocular}, indicating the reliability of our depth predictions. Since Cui et al.’s method was trained on the UWA Database where the ground truth is not real depth, we do not include depth error comparisons on this dataset for fairness.

\begin{table}[!t]
% \renewcommand{\arraystretch}{1.3}
% if using array.sty, it might be a good idea to tweak the value of
% \extrarowheight as needed to properly center the text within the cells
\caption{Weighted depth error (mm) compared with Cui et al \cite{cui2023monocular}.}
\label{depth}
\centering
% Some packages, such as MDW tools, offer better commands for making tables
% than the plain LaTeX2e tabular which is used here.
\begin{tabular}{c c c c c}
\toprule
\textbf{Dataset} & \textbf{Cui et al. \cite{cui2023monocular}}  & \textbf{Proposed}\\
\midrule
PolyU 3D+ & 2.9577 & \textbf{1.7876}\\
\bottomrule
\end{tabular}
\vspace{-10pt}
\end{table}

\subsection{3D Fingerprint Registration Accuracy}

We verify that our method can correctly register contactless fingerprints. Previous 2D registration methods \cite{bazen2003matching}\cite{si2015detection}\cite{cui2020dense}\cite{guan2024phase}\cite{jia2025single} cannot obtain 3D information of contactless fingerprints, and thus can only register contactless fingerprints at the 2D image level. Our method, however, can effectively register them in 3D space by predicting the depth of contactless fingerprints in 3D space, thereby significantly improving the accuracy of contactless fingerprint registration. In the registration experiment, only the results of the registered fingerprint point cloud are used for experiments, without involving the subsequent new view rendering and synthesis.

Previous fingerprint registration methods often use the correlation coefficient of binarized fingerprints as a metric for registration accuracy. However, since the binarization of contactless fingerprints is often inaccurate, we adopt the average distance of minutiae points in the test set after registration as the metric for registration accuracy. That is, given the pixel coordinates $(\textbf{x},\textbf{y})$ and mated $(\textbf{x}',\textbf{y}')$ of minutiae point pairs in two images, their pixel distance is

$$
D = \frac{1}{N} \sum_{i=1}^{N} \sqrt{(\textbf{x}_i - \textbf{x}_i')^2 + (\textbf{y}_i - \textbf{y}_i')^2 }
$$

 We use Verifinger 13.0 \cite{verifinger} to extract minutiae points and match minutiae relationships on the UWA dataset, and manually adjust the minutiae points and matching relationships through manual adjustment to obtain more precise ground truth label. Considering the large of manual annotation, we randomly annotate minutiae points and matching relationships for only 46 fingers, and perform matching between them, obtaining 6 pairs of matches for each finger. 

\begin{figure}[!bt]
\centering
\centerline{\includegraphics[width=\linewidth]{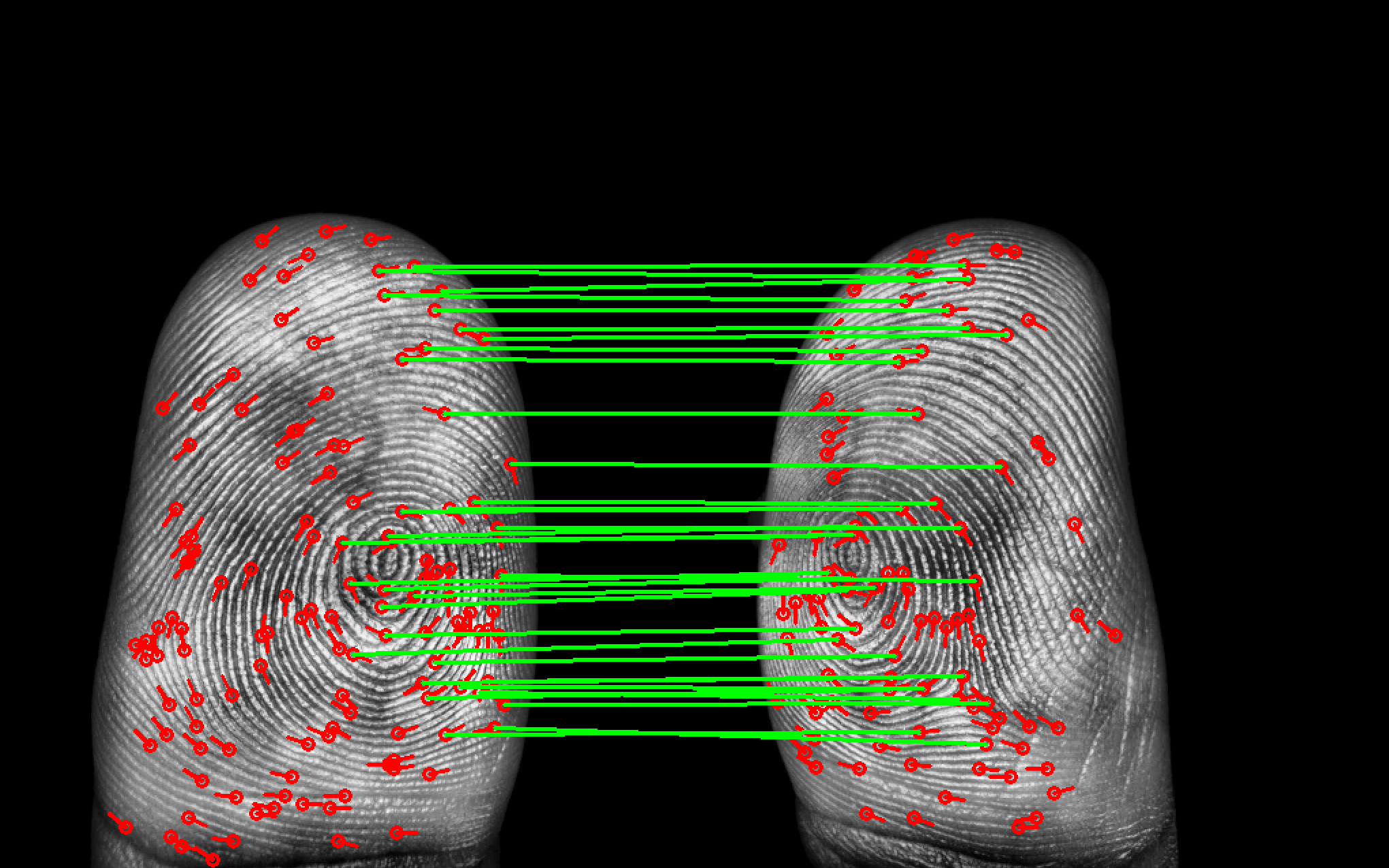}}
\caption{Manually annotate ground truth minutiae points and matching relationships.}
\label{fig:maker}
\vspace{-5pt}
\end{figure}

\begin{figure*}[!bt]
\centering
\centerline{\includegraphics[width=0.7\linewidth]{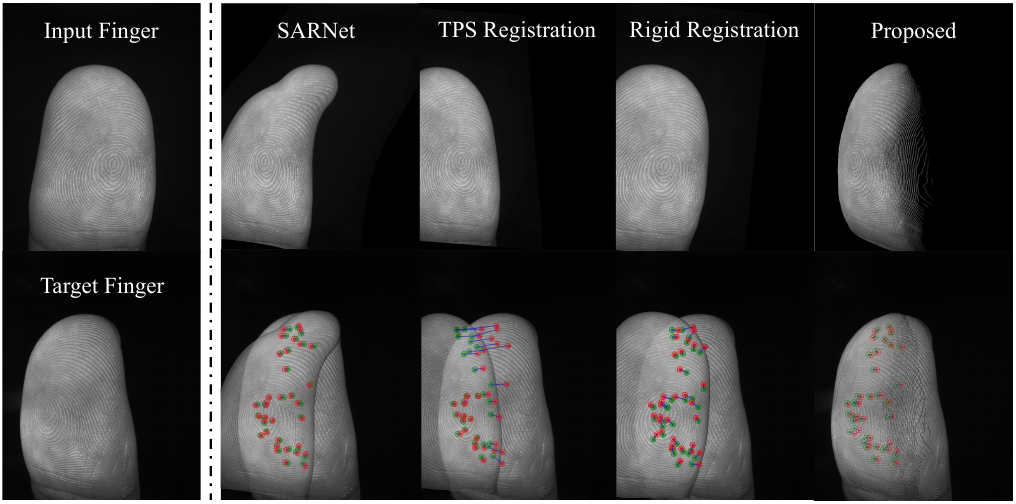}}
\caption{Examples of 3D registration results of a pair of contactless fingerprints with a 45-degree difference, showing that 2D methods cannot handle fingerprint rotation effectively in three-dimensional space, and may even produce physically implausible distortions; meanwhile, our method can effectively handle contactless fingerprint registration with large angle difference, and remains physically plausible, while other methods cannot produce reasonable images.}
\label{fig:regist}
\end{figure*}

\begin{figure}[!bt]
\centering
\centerline{\includegraphics[width=\linewidth]{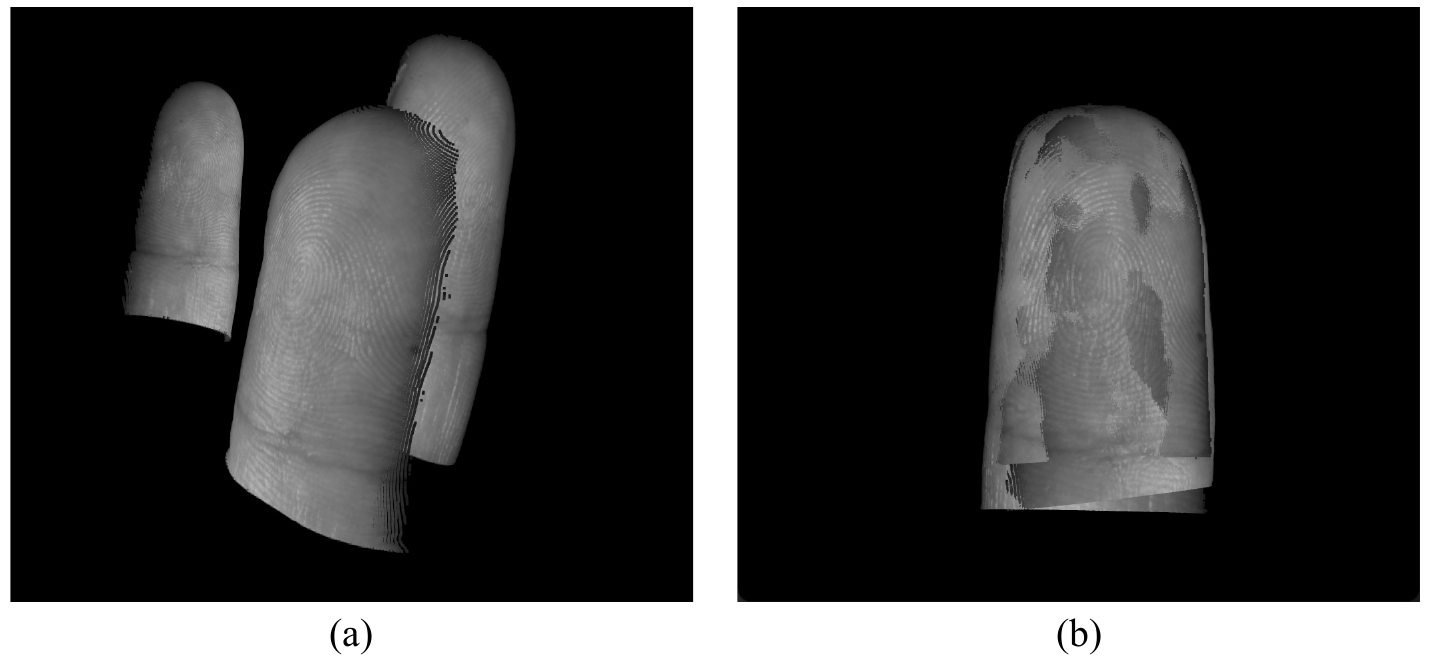}}
\caption{Registration comparison between VGGT and our method on CFPose. VGGT often fails to register completely. (a) VGGT, (b) Proposed.}
\label{fig:vggt_cfpose}
\vspace{-10pt}
\end{figure}

Since fingerprint textures with a 90-degree pose difference almost have no intersection, which are meaningless for evaluation, only fingerprints with a 45-degree pose difference are selected for testing, resulting in 46*4=184 registration pairs. Additionally, considering that manually annotated test sets might be suspected of cherry-picking good results, we also test registration accuracy on 6214 pairs from 1167 fingers where Verifinger could find matching minutiae points. However, due to the large angle difference between left and right views, Verifinger often makes incorrect minutiae matching, resulting in poor matching performance. We ultimately remove matches with a 90-degree pose difference and conduct registration experiments on the remaining 4312 pairs. We also conduct similar experiments on CFPose. As shown in Table \ref{table:regist}, for contactless fingerprint datasets with large pose spans like UWA, 3D registration methods have a clear advantage over 2D registration. However, for datasets like CFPose with smaller pose differences, 2D registration methods still have certain advantages, although the state-of-the-art 3D registration method VGGT \cite{wang2025vggt} performs poorly on CFPose, our method can still achieve good results.

% Our 3D registration is achieved by projecting minutiae points from one image back to the 3D world space and then rendering from the camera position of another image.
\begin{table}

\caption{Comparisons of registration accuracy across different methods and datasets}
\centering
\begin{tabular}{c|c|ccc}
\midrule
\multicolumn{2}{c|}{\diagbox{\textbf{Methods}}{\textbf{Test Set}}} & \textbf{UWA} & \textbf{UWA}*  & \textbf{CFPose}* \\
\midrule
\multirow{3}{*}{2D} & Rigid Transform & 52.27 & 73.47 & 42.58 \\
& TPS Based \cite{bazen2003matching} & 26.18 & 6.83$^{1}$ & 4.67$^{1}$ \\
& SARNet \cite{jia2025single} & 16.86 & 39.09 & \textbf{25.57} \\
\midrule
\multirow{2}{*}{3D} & VGGT \cite{wang2025vggt} & 10.45$^{2}$ & 28.69$^{2}$ & 303.0$^{2}$ \\
& Proposed & \textbf{13.94} & \textbf{30.24} & 40.07 \\
\bottomrule
\end{tabular}

\begin{tablenotes}
\item {*}: Represents the minutiae matching relationships predicted by Verifinger, which are not necessarily correct.
\item {1}: The TPS uses the predicted minutiae pairs to conduct deformation, with the ground truth also being the predicted minutiae pairs, thus the minutiae errors are unrealistically low. Using TPS aligns the minutiae pairs as much as possible without considering image plausibility, which leads to artificially low errors.
\item {2}: Although VGGT performs well on UWA, it completely fails to work on CFPose.
\end{tablenotes}
\label{table:regist}
\vspace{-5pt}

\end{table}

\begin{figure*}[htb]
\begin{center}
% \subfloat %子图片标题
% {\includegraphics[width=1\linewidth]{imgs/example_matching.pdf}}

% \centerline{\includegraphics[width=\linewidth]{imgs/uwa_det3.eps}} %[图片大小]{图片路径}
\subfloat{\includegraphics[width=0.45\linewidth]{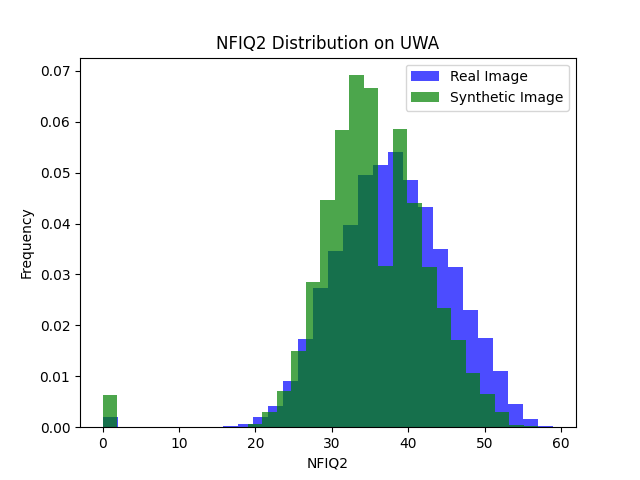}}
\subfloat{\includegraphics[width=0.45\linewidth]{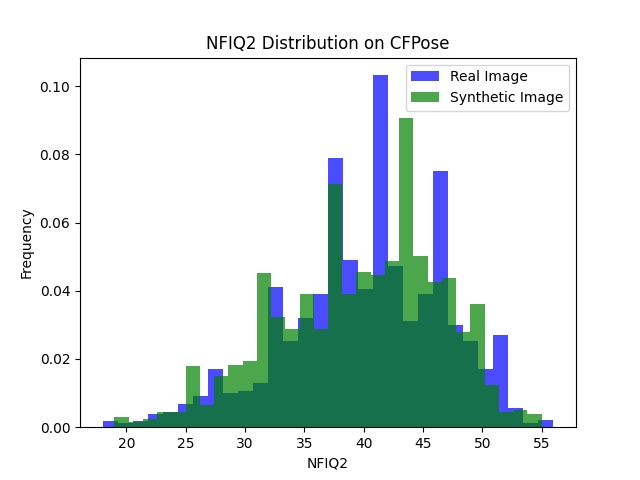}}
\end{center}
% \vspace{-10pt} % 减少图表下方的间距

% \caption{DET curves on UWA, CFPose and ZJU databases} %图片标题
\caption{NFIQ2 distributions on UWA (left) and CFPose (right) databases of real images and our synthetic images} %图片标题

\label{fig:nfiq_dis}  %图片交叉引用时的标签
\vspace{-10pt} % 减少图表下方的间距

\end{figure*}
The experimental results show that 3D registration methods have clear advantages over 2D registration methods on datasets with large pose variations captured from different angles, such as UWA. We visualize the registration results of several methods for contactless fingerprints in Fig. \ref{fig:regist}, revealing that 2D methods produce registrations with obvious physically implausible deformations, while 3D registration methods can correctly register contactless fingerprints in accordance with physical laws. We do not conduct comparative experiments with dense registration methods \cite{si2015detection}\cite{cui2020dense}\cite{guan2024phase} because these methods are based on TPS-based registration, and when TPS-based methods fail, these methods also fail, yielding results similar to TPS-based methods, making such experiments unnecessary. Since it is difficult to completely annotate the correct matching relationships of minutiae points manually, we only annotated the minutiae points that can be obviously observed manually when annotating the matching relationships, which is why there aren't many matching minutiae points in Fig. \ref{fig:regist}.

Although the recently proposed VGGT\cite{wang2025vggt} shows better performance on the UWA dataset, the method completely fails to handle the CFPose dataset as shown in Fig. \ref{fig:vggt_cfpose}. This is mainly because VGGT's training set consists entirely of static objects. UWA captures fingers from three angles simultaneously, which is consistent with VGGT's training set construction method. However, CFPose uses fixed camera positions to capture fingers in different poses, which doesn't align with VGGT's training set, making it difficult to work effectively. Additionally, since VGGT is an end-to-end framework, it's challenging to adapt it for CFPose.

\subsection{Synthesized Fingerprint Data Quality}

We evaluate the quality of the synthesized images using NFIQ2 for fingerprint quality assessment by Verifinger. For real images, UWA use nearly 4,500 images from the three angles of the first capture to calculate the average NFIQ2, with missing images assigned a NFIQ2 of 0; CFPose calculate the average NFIQ2 for all images. As shown in Table \ref{NFIQ2}, the quality of our generated contactless fingerprints is relatively close to that of the real datasets. 
% Additionally, we can see that introducing SAGD segmentation effectively improves the quality of the synthesized fingerprint images.
Since VGGT fails to perform effective registration on CFPose, we do not evaluate its NFIQ2. Fig. \ref{fig:nfiq_dis} shows the NFIQ2 distributions of the contactless fingerprint images synthesized by our method and the real fingerprint images. It can be observed that the distributions are relatively close on both datasets, indicating that our method can effectively synthesize contactless fingerprint data.
\begin{table}[!t]
% \renewcommand{\arraystretch}{1.3}
% if using array.sty, it might be a good idea to tweak the value of
% \extrarowheight as needed to properly center the text within the cells
\caption{NFIQ2 on both real and synthetic contactless fingerprints.}
\label{NFIQ2}
\centering
% Some packages, such as MDW tools, offer better commands for making tables
% than the plain LaTeX2e tabular which is used here.
\begin{tabular}{c |c | c c c}
\toprule
\textbf{Dataset} & \textbf{Real image}  & \textbf{Mast3R \cite{leroy2024grounding}} & \textbf{VGGT \cite{wang2025vggt}}  & \textbf{Proposed}\\
\midrule
UWA & 38.13  & 35.48 & 35.46 & \textbf{35.68}\\
\midrule
CFPose & 40.19  & 38.43 & - & \textbf{39.57}\\
\bottomrule
\end{tabular}
\end{table}

\begin{table}[!t]
% \renewcommand{\arraystretch}{1.3}
% if using array.sty, it might be a good idea to tweak the value of
% \extrarowheight as needed to properly center the text within the cells
\caption{Matching performances between synthetic and real fingerprints using Verifinger on UWA database.}
\label{matching}
\centering
% Some packages, such as MDW tools, offer better commands for making tables
% than the plain LaTeX2e tabular which is used here.
\begin{tabular}{c c c c c}
\toprule
\textbf{Dataset} & \textbf{shot} & \textbf{EER} & \textbf{FMR@1\%} & \textbf{FMR\_Zero}\\
\midrule
Real Image$^1$ & 1\&2 & \textit{32.30\%} & \textit{62.87\%} & \textit{69.41\%}\\
\midrule
\multirow{2}{*}{InstantSplat$^1$} & 1 & 31.79\% & 61.68\% & 66.59\%\\
~ & 2 & 34.17\% & 64.27\% & 70.38\%\\

\midrule
\multirow{2}{*}{Proposed$^1$}& 1 & \textbf{28.00\%} & \textbf{61.47\%}& \textbf{66.50\%}\\
~ & 2 &  \textbf{30.00\%} & \textbf{64.01\%}& \textbf{70.22\%}\\
\toprule
\bottomrule

Real Image$^2$ & 1\&2 & \textit{31.48\%} & \textit{63.40\%} & \textit{69.84\%}\\
\midrule
\multirow{2}{*}{InstantSplat$^2$} & 1 & 32.48\% & 65.24\% & 74.85\%\\
~ & 2 & 34.75\% & 69.28\% & 79.26\%\\

\midrule
\multirow{2}{*}{Proposed$^2$}& 1 & \textbf{26.05\%} & \textbf{64.64\%}& \textbf{74.25\%}\\
~ & 2 &  \textbf{27.17\%} & \textbf{68.84\%}& \textbf{78.80\%}\\
\bottomrule
\end{tabular}

\begin{tablenotes}
\item {1}: Following \cite{cui2023monocular}, the matching test is conducted with both real and synthetic images sized at 3000*3000.
\item {2}: For the matching test on the entire dataset, the real images are 4500*4500 and the synthetic images are 18000*4500.
\end{tablenotes}
\vspace{-10pt}
\end{table}

\begin{table}[!t]
% \renewcommand{\arraystretch}{1.3}
% if using array.sty, it might be a good idea to tweak the value of
% \extrarowheight as needed to properly center the text within the cells
\caption{Matching performances between synthetic and real fingerprints using Verifinger on CFPose database.}
\label{matching_cfpose}
\centering
% Some packages, such as MDW tools, offer better commands for making tables
% than the plain LaTeX2e tabular which is used here.
\begin{tabular}{c c c c c}
\toprule
\textbf{Dataset} & \textbf{EER} & \textbf{FMR@1\%} & \textbf{FMR\_Zero}\\
\midrule
Real Image & \textit{11.38\%} & \textit{16.38\%} & \textit{31.35\%}\\
\midrule

InstantSplat &  15.98\% & 29.15\% & 49.18\%\\
% \midrule
% Proposed$^\dag$ &  15.11\% & 26.98\% & 46.38\%\\
\midrule
Proposed &  \textbf{13.40\%} & \textbf{24.90\%} & \textbf{44.47\%}\\
\bottomrule
\end{tabular}
\end{table}

\begin{figure*}[!bt]
\centering
\centerline{\includegraphics[width=\linewidth]{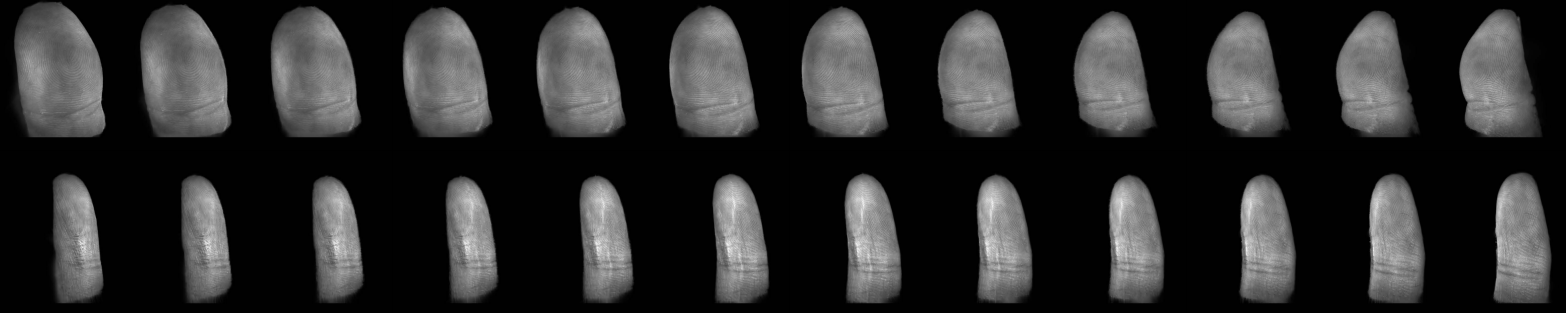}}
\caption{Examples of synthetic contactless fingerprints of 12 different views.}
\label{fig:visulization}
ji\end{figure*}

\begin{table*}[t]
\centering
\caption{Matching Performances of Synthetic Fingerprints at Different Angles}

\begin{tabular}{cc}
\begin{tabular}{ccccc}
\multicolumn{5}{c}{\textbf{Matched with Center Pose}} \\
\midrule
Angle Diff ($^\circ$) & EER & FMR@1\% & FMR@0.01\% & FMR@0\%\\
\midrule
-45    & 0.5287 & 0.7367 & 0.9487 & 0.9553 \\
-36.82 & 0.4393 & 0.6213 & 0.8947 & 0.9113 \\
-28.67 & 0.3353 & 0.5193 & 0.8207 & 0.8453 \\
-20.45 & 0.1747 & 0.2453 & 0.4780 & 0.5047 \\
-12.27 & 0.1147 & 0.1273 & 0.1740 & 0.1820 \\
-4.091 & 0.0813 & 0.0847 & 0.0933 & 0.0953 \\
4.091  & 0.0787 & 0.0813 & 0.0873 & 0.0873 \\
12.27  & 0.1127 & 0.1220 & 0.1540 & 0.1600 \\
20.45  & 0.1660 & 0.2093 & 0.3293 & 0.3487 \\
28.67  & 0.3080 & 0.4573 & 0.7527 & 0.7693 \\
36.82  & 0.4507 & 0.6387 & 0.8913 & 0.9060 \\
45     & 0.5713 & 0.7460 & 0.9407 & 0.9507 \\
\midrule
\end{tabular} &
\begin{tabular}{ccccc}
\multicolumn{5}{c}{\textbf{Matched with Left Pose}} \\
\midrule
Angle Diff ($^\circ$)& EER & FMR@1\% & FMR@0.01\% & FMR@0\%\\
\midrule
0      & 0.0678 & 0.0760 & 0.0860 & 0.0860 \\
8.18   & 0.0893 & 0.1180 & 0.1647 & 0.1673 \\
16.36  & 0.1387 & 0.3460 & 0.5540 & 0.5713 \\
24.55  & 0.2024 & 0.6227 & 0.8633 & 0.8747 \\
32.73  & 0.2229 & 0.6653 & 0.8967 & 0.9180 \\
40.91  & 0.2367 & 0.7133 & 0.9380 & 0.9487 \\
49.09  & 0.2833 & 0.8093 & 0.9740 & 0.9807 \\
57.27  & 0.4153 & 0.9033 & 0.9913 & 0.9920 \\
65.45  & 0.5587 & 0.9640 & 0.9973 & 0.9980 \\
73.64  & 0.6547 & 0.9813 & 0.9973 & 0.9980 \\
81.82  & 0.6647 & 0.9793 & 0.9980 & 0.9993 \\
90.0   & 0.6813 & 0.9847 & 1.0000 & 1.0000 \\
\midrule
\end{tabular} \\
\end{tabular}
\label{tab:angle_diff}
\vspace{-10pt}
\end{table*}

\begin{figure*}[htb]
\begin{center}
% \subfloat %子图片标题
% {\includegraphics[width=1\linewidth]{imgs/example_matching.pdf}}

% \centerline{\includegraphics[width=\linewidth]{imgs/uwa_det3.eps}} %[图片大小]{图片路径}
\subfloat{\includegraphics[width=0.45\linewidth]{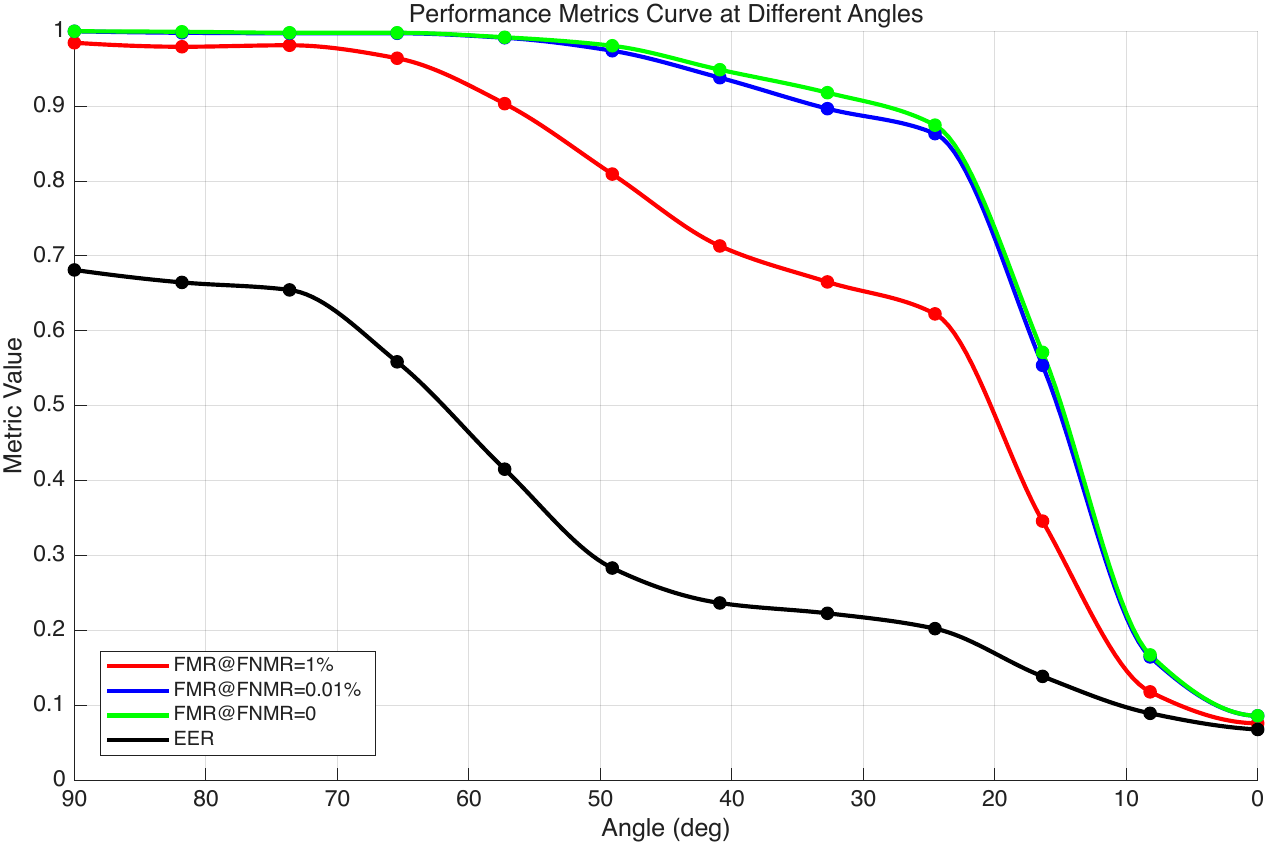}}
\subfloat{\includegraphics[width=0.45\linewidth]{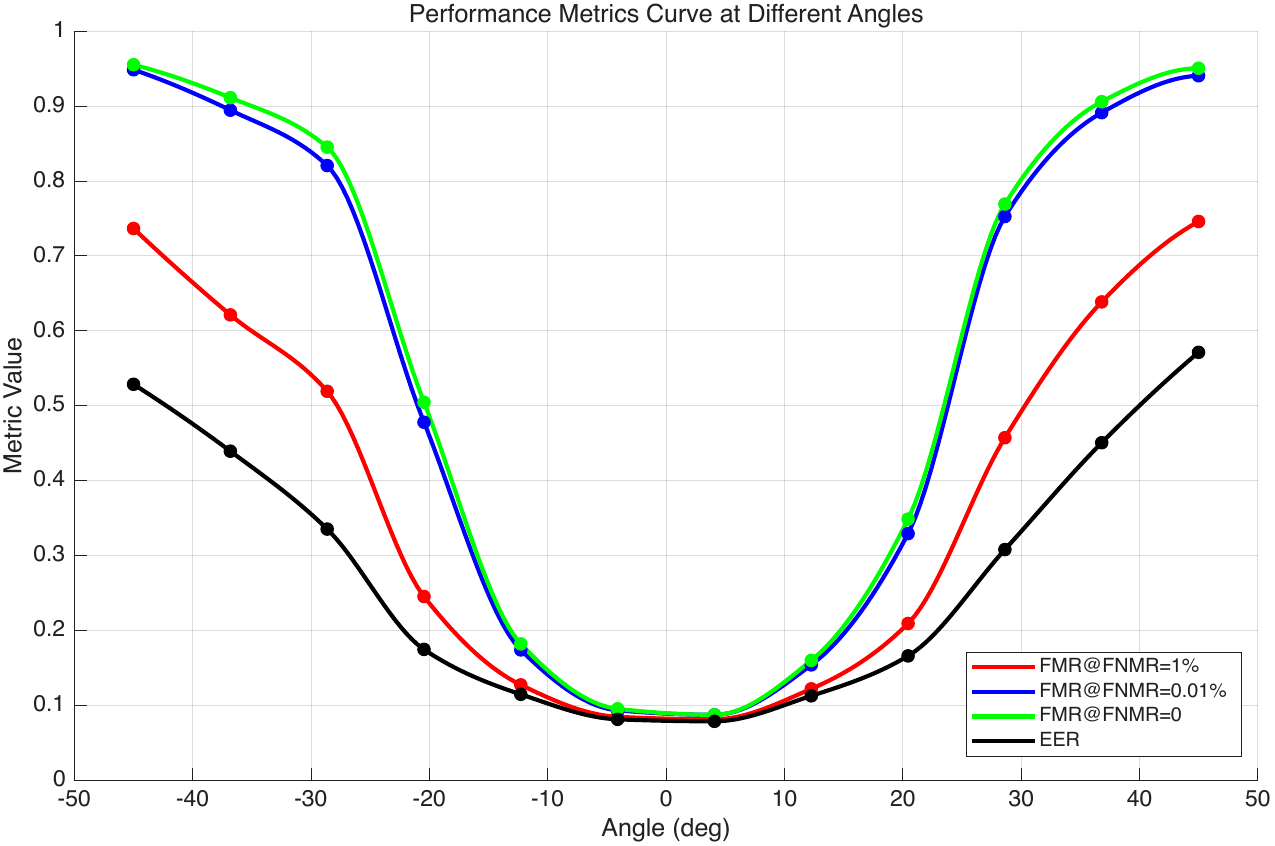}}
\end{center}
% \vspace{-10pt} % 减少图表下方的间距

% \caption{DET curves on UWA, CFPose and ZJU databases} %图片标题
\caption{Performance for synthetic fingerprint matching across different angles. Left from $90^{\circ}$ to $0^{\circ}$. Right from $-45^{\circ}$ to $45^{\circ}$} %图片标题

\label{fig:angle_diff}  %图片交叉引用时的标签
% \vspace{-10pt} % 减少图表下方的间距

\end{figure*}

\subsection{Synthesized Fingerprints for Fingerprint Matching}

Subsequently, we match synthesized images to real images for testing. For the UWA dataset, there are total of $1500\times 12 = 18000$ synthesized images, and real images are $1500\times 3 = 4500$ images each time. Missing images are assigned a matching score of 0. Following previous matching principles \cite{tan2020towards}\cite{dong2023syn}\cite{cui2023monocular} on UWA, we match 3000 images from the first 100 fingers, comparing real images from the first and second captures, synthesized images with real images from the first capture, and synthesized images with real images from the second capture. Each comparison involves $3000 \times 3000$ matches, resulting in 9000 genuine matches and 8,991,000 impostor matches. The synthesized images selected are the 1st, 7th, and 12th from 12 views. As shown in Table \ref{matching}, the results of synthesized images are very close to the matching results between real images. The three angles from the first capture are used to synthesize 3D fingerprints, and the matching performance of synthesized fingerprint images are somewhat better when matched with the first capture compared to the second capture. We use the original InstantSplat \cite{fan2024instantsplat} pipeline without any modifications as baseline method.

Subsequently, since our method does not involve a training set, we performed matching on the entire dataset. We conduct $18000 \times 4500$ matching experiments for both the first and second captures, resulting in 54,000 genuine matches and 80,946,000 impostor matches. It can be seen that the matching results between synthesized data and real data are also very close.

It is worth mentioning that our results using Verifinger matching are not as good as previous outcomes \cite{cui2023monocular}\cite{tan2020towards}\cite{dong2023syn}. This is because we use the original fingerprint images rather than preprocessed ones (histogram equalization, rotation correction, frequency normalization, etc.). Whether fingerprints are preprocessed or not does not affect our demonstration that the proposed method can synthesize contactless fingerprints with quality comparable to the original images.

We also conduct similar experiments on CFPose. Specifically, the real image matching involved 1400*1399/2 = 979,300 total pairs, including 6,300 genuine matches and 973,000 impostor pairs. For synthetic images, the matching involve 140*12*140*10 = 2,352,000 pairs, including 16,800 genuine matches and 2,335,200 impostor pairs. The experimental results demonstrate that our performance closely approaches that on real images, and our method effectively improves the matching performance.

% \subsection{The Impact of Viewpoint on Contactless Fingerprint Matching}

We further detailedly test the synthesized fingerprint images with different pose angles. In the UWA dataset, each fingerprint is captured from the left, front, and right perspectives. But our method is able to synthesize any view angle of contactless fingerprints using images from three angles. In our experiment, we generate 12 different views in reality, and test those different views' matching performances. We match the 12 synthesized new fingerprints with the real images from the left and front views. We consider the left angle to be -45 degrees, the front to be 0 degrees, and the right to be 45 degrees, and perform 1500*1500 matching for each of the 12 angles. The final results, as shown in Table \ref{tab:angle_diff} and Fig. \ref{fig:angle_diff},  indicate that contactless fingerprint matching performs well when the angle difference is less than 20 degrees, but matching performance deteriorates as the angle difference increases. When the angle difference reaches 90 degrees, the performance becomes very poor.

Fig. \ref{fig:visulization} qualitatively illustrates some synthesized results with different poses, demonstrating that our method can generate high-quality contactless fingerprint images with large pose variations.

\subsection{Synthesized data for Contactless Fingerprint Recognition}

We further test our synthesized for training deep neural networks to exam its improvement for the recognition performance of contactless fingerprints. In this section, we use the open-source DeepPrint \cite{rohwedder2023benchmarking} for training contactless fingerprint recognition network. We use the last 50 fingers of the UWA dataset as the training set, training with both original images (about 3000) and original images + synthesized images (about 3000+18000), and test on the first 100 fingers of UWA (about 6000 images) and the first 120 fingers of CFPose (about 1200 images). Moreover, we do not include any real or synthetic data from CFPose in our training set; however, we conduct matching experiments on the first 120 fingers of CFPose.

We use Verifinger to extract contactless fingerprint minutiae as ground truth for the Minutiae branch. To avoid potential inaccuracies in the minutiae extracted by Verifinger, we train both a DeepPrint with only the Texture branch and a DeepPrint with both Texture and Minutiae branches. All trainings are conducted for 100 epochs. As shown in Table \ref{deepprint} and \ref{deepprint_cfpose} and Fig. \ref{deepprint}, after adding our synthesized fingerprints, the matching performances of networks trained using both methods improved significantly.

\begin{table}[!t]
% \renewcommand{\arraystretch}{1.3}
% if using array.sty, it might be a good idea to tweak the value of
% \extrarowheight as needed to properly center the text within the cells
\caption{Matching performances between synthetic and real fingerprints by DeepPrint on UWA database.}
\label{deepprint}
\centering
% Some packages, such as MDW tools, offer better commands for making tables
% than the plain LaTeX2e tabular which is used here.
\begin{tabular}{c c c c c}
\toprule
\textbf{Dataset} & \textbf{shot} & \textbf{EER} & \textbf{FMR@1\%} & \textbf{FMR\_Zero}\\
\midrule
\multirow{2}{*}{DeepPrint \cite{rohwedder2023benchmarking}} & Tex & 31.70\% &	72.56\% & 90.24\% \\
~ & Tex + Minu& 28.66\% &	66.90\%	& 86.62\%\\

\midrule
\multirow{2}{*}{Proposed}& Tex & \textbf{28.55\%} & \textbf{67.06\%}& \textbf{83.69\%}\\
~ & Tex + Minu &  \textbf{26.37\%} & \textbf{63.46\%}& \textbf{83.41\%}\\

\bottomrule
\end{tabular}
\vspace{-10pt}

\end{table}

\begin{table}[!t]
% \renewcommand{\arraystretch}{1.3}
% if using array.sty, it might be a good idea to tweak the value of
% \extrarowheight as needed to properly center the text within the cells
\caption{Matching performances between synthetic and real fingerprints by DeepPrint on CFPose database.}
\label{deepprint_cfpose}
\centering
% Some packages, such as MDW tools, offer better commands for making tables
% than the plain LaTeX2e tabular which is used here.
\begin{tabular}{c c c c c}
\toprule
\textbf{Dataset} & \textbf{shot} & \textbf{EER} & \textbf{FMR@1\%} & \textbf{FMR\_Zero}\\
\midrule
\multirow{2}{*}{DeepPrint \cite{rohwedder2023benchmarking}} & Tex & 5.67\% & 24.59\% & 75.58\% \\
~ & Tex + Minu& 5.56\% & \textbf{16.85\%}  & 63.08\% \\

\midrule
\multirow{2}{*}{Proposed}& Tex &  \textbf{5.52\%} & \textbf{21.54\%} & \textbf{70.87\%}\\
~ & Tex + Minue &  \textbf{5.53\%} & 17.61\% & \textbf{60.15\%}\\

\bottomrule
\end{tabular}
\vspace{-10pt}

\end{table}

\begin{figure}[htb]
\begin{center}
% \subfloat %子图片标题
% {\includegraphics[width=1\linewidth]{imgs/example_matching.pdf}}

% \centerline{\includegraphics[width=\linewidth]{imgs/uwa_det3.eps}} %[图片大小]{图片路径}
\subfloat{\includegraphics[width=0.9\linewidth]{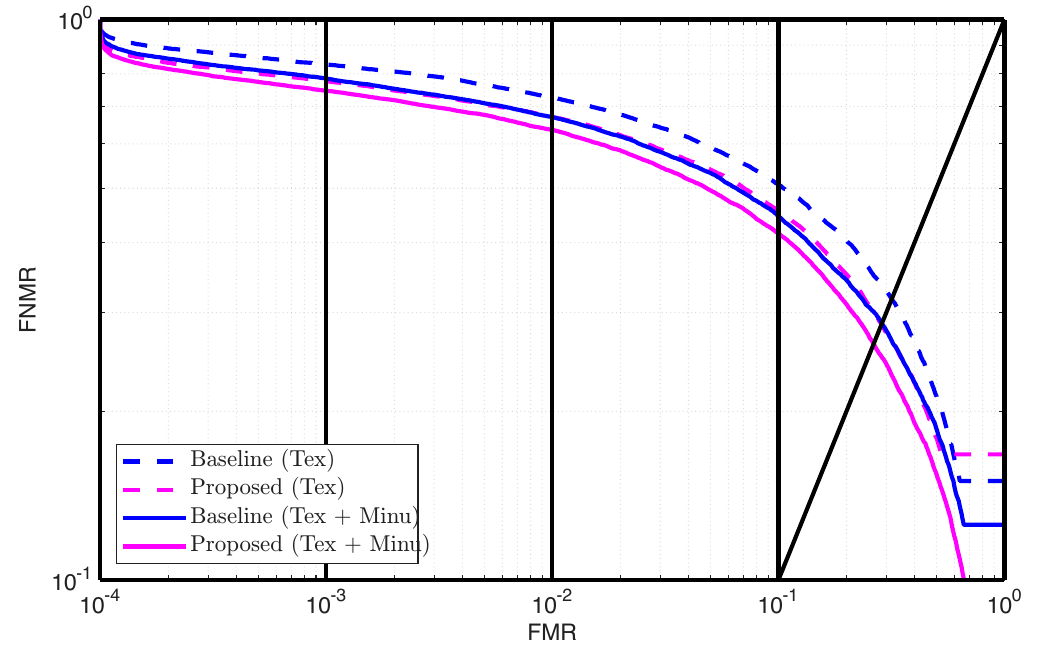}}
\end{center}
% \vspace{-10pt} % 减少图表下方的间距

% \caption{DET curves on UWA, CFPose and ZJU databases} %图片标题
\caption{DET curves for training a fingerprint matching network with or without our synthetic data} %图片标题

\label{fig:deepprint}  %图片交叉引用时的标签
% \vspace{-10pt} % 减少图表下方的间距
\vspace{-10pt}

\end{figure}

\section{Conclusion}
In this research, we innovatively propose a framework for contactless fingerprint 3D reconstruction, registration, and fingerprint generation. It is the first method to introduce 3D Gaussian Splatting (3D-GS) into the fingerprint recognition field, achieving high-quality 3D fingerprint synthesis on public contactless fingerprint datasets.
Moreover, we also propose a 3D fingerprint registration method, which is more physically consistent and interpretable compared to traditional 2D registration methods when applied to contactless fingerprints. Most importantly, the newly synthesized contactless fingerprint can be used in training fingerprint recognition neural networks, effectively improving the performance of contactless fingerprint recognition.

% or


% \appendices

% % you can choose not to have a title for an appendix
% % if you want by leaving the argument blank
% \section{}

\section*{Acknowledgment}

This work is supported in part by the National Natural Science Foundation of China under Grants 62206026.

\ifCLASSOPTIONcaptionsoff
  \newpage
\fi



%\bibliographystyle{IEEEtran}



% \end{thebibliography}
\bibliographystyle{IEEEtran}
\bibliography{egbib}

% \end{IEEEbiography}


% % insert where needed to balance the two columns on the last page with
% %\newpage



%\vfill


\end{document}